\documentclass{article}

\usepackage{arxiv}

\usepackage[utf8]{inputenc} 
\usepackage[T1]{fontenc}    
\usepackage{hyperref}       
\usepackage{url}            
\usepackage{booktabs}       
\usepackage{amsfonts}       
\usepackage{nicefrac}       
\usepackage{microtype}      
\usepackage{lipsum}
\usepackage{multirow}
\usepackage{graphicx}
\usepackage{subfig}
\usepackage{setspace} 
\title{Adaptive Neuro Particle Swarm Optimization applied for diagnosing disorders}

\author{
  Majid Masoumi \\
  Department of Computer Science\\
  Yazd University\\
  \texttt{majid.masoumi@iau.yazd.edu.ir} \\
   \And
 Mina Rajabi  \\
  Department of Computer Science\\
  Yazd University\\
    \texttt{rajabi.mina@iau.yazd.edu.ir} \\
    \And
}
\begin{document}
\maketitle

\begin{abstract}
A new Adaptive Neuro Particle Swarm Optimization (ANPSO) combined with a fuzzy inference system for diagnosing disorders is presented in this paper. The main contributions of the novel proposed method can be a global search across the whole search space with faster convergence rate. Moreover, it shows a better exploration and exploitation by applying the adaptive control parameters, automatic control of inertia weight and coefficient of personal and social behaviours. Utilizing such attributes lead to a fast and smart diagnosis mechanism which is able to diagnosis the diseases by the high accuracy. The ANPSO is associated with tuning the characteristics of the inference system to achieve the minimum diagnosis error as far as the optimized model is obtained. As a case study, we use liver disorders dataset called Bupa. According to the preliminary ramifications, the suggested adaptive PSO performance can overcome the traditional inference system and combined with other optimization methods substantially.

\end{abstract}

\keywords{Adaptive \and Fuzzy Inference Systems \and Nuro Inference System \and Optimization\and Adaptive Particle Swarm Optimization \and Diagnosing disorders.  }

\section{Introduction}
Medical inference systems are a category of Artificial Intelligence (AI) application that supports the free practice of techno-scientific human knowledge for solving semi or ill-structured matters wheresoever there is not a special promise for defining the algorithm. The Medical inference systems have been characterised as intelligent employment which appropriates knowledge and inference levels to handle severe problems to necessitate significant human expertise for their clarifications \cite{feigenbaum1981expert}.  

A fuzzy inference system is a particular knowledge-based system, which is made of fuzzification, knowledge database, inference rules,  and defuzzification parts, and uses fuzzy sets instead of the Boolean logic to hold applied data in the deduction mechanism. This system is utilised to explain decision-making problems, where there is no scientific and straightforward method. However,  the problem solution can be studied alternatively, which is in regard to professionals in the form of If-Then rules. A fuzzy inference system can be adequately provided to the problem, which gives uncertainty emitting from fuzziness, ambiguity or subjectivity. In the 21st century, the attentions of FIS have been tremendously growing wherever scientific research issues such as diagnosing and predicting the different risks of the diseases \cite{polat2006diagnosis,csahan2007new,adeli2010fuzzy,neshat2015new,moya2019fuzzy}, civil engineering applicabilities as an assistant because a pretty high safety level is needed in civil engineering compositions the uncertainty associated with the relationship of scientific estimations are very roughly and conservatively calculated handling traditional methods\cite{neshat2011designing, neshat2012predication,yuan2014prediction,chiew2017fuzzy}, assessing the educational service qualities by embedded fuzzy sets\cite{pourahmad2012service,pourahmad2016using, raeesi2018quality,du2018fuzzy,shafii2016assessment} and in representing various aspects of indeterministic business positions \cite{neshat2011fhesmm,neshat2016designing,rudzewicz2015quality,deveci2018interval}.  

Recently, Efforts have been performed to develop the PSO performance, and also a few PSO modifications have been introduced. Much practice centred on parameters settings of the algorithm \cite{shi1998modified} and on combining various techniques into the PSO \cite{ratnaweera2004self, angeline1998using, brits2002niching}. However, most of these developed PSOs manage the control parameters or hybrid operators outdoors regarding the varying states of evolution. Therefore, these actions have a lack of well-organised strategy of evolutionary state and besides seldom experience from an insufficiency in dealing with complex search spaces. 

 Consequently, we have investigated to propose an adaptive mechanism for updating the control parameters of PSO in real-time mode. In this way, the main contributions are including 

1.Applying a robust evolutionary idea called 1+1EA with an adaptive mutation step size for optimizing and tuning the control paraMeters of PSO.

2.Using the ANFIS because of its benefits such as simple sense, immense flexibility, the capacity to endure erroneous data properly, creating complex non- linear functions, to act on the basis of skilful knowledge, the ability to comply with traditional controlling systems.

3. Applying the proposed adaptive PSO to develop the performance of diagnosis by tuning the hyper-parameters of ANFIS like the number and kind of fuzzy membership functions and heightening the fuzzy rules.    

The main goal of this article is to address the important benefits and shortcomings of the current approaches and theories for improving and modelling fuzzy inference systems compared with the proposed adaptive neuro particle swarm optimization performance for diagnosing the liver disorders based on the dataset. Concerning accomplishing like plans, a comprehensive study of the relevant fuzzy inference system technique is utilized.

The rest of this article is arranged as views. The details of the dataset used can be seen in Section 2. Section 3 illustrates the fuzzy inference systems in details.  Besides, the technical specifications of the particle swarm optimization method and its diversity are studied in Section 4, and also section 6 shows the experimental implications. Conclusively, conclusions are sketched in Section 5.

\section{Liver Disorders and Data-Set}
The liver is one of the most important body glands that detoxification of prescriptions, removal of natural things emerging from the demolition and rehabilitation of RBCs in the form of bile, composition of blood clotting parts, area of sugar as glycogen. Furthermore, the ordinance of sugar and fat metabolism are some of the fundamental functions of this body organ. We must not underestimate its function in fat metabolism and defence before microbus and toxins coming from foodstuff \cite{farokhzad2016novel}. In the last decade, the death damage following different liver disorders has been dramatically growing. On-time examination of this disease can be useful in the inhibition of its effects, its control and treatment. Browse contemplates expert's mentality as one of the most critical issues in diagnosing disease because human-being is subject to error, and there is a possible error in disease diagnosis. One of the notable informatics medical procedures is to use expert systems to diagnose the disease concerning a group of symptoms. These schemes can be based on artificial intelligence (AI) and assist experts to diagnose the diseases and more adequately satisfy them by acknowledging laboratory examinations. They also decrease cost,  save the time of experts and their incorrect judgment.
The dataset employed in this paper, which is used for enhancing the ability of liver disorders investigation according to their qualities, gathered by Richards forsyth and presented to the UCI in 1995. The samples number in this collection are 345, and each sample consists of 7 attributes.  In this dataset the first five fields are reported to variable substances of a male blood test, the 6th field is the quantity of
alcohol drinking, and lastly, the 7th field is using for restricting the healthy or ill individual. Characteristic information can be seen in the following:\\
1.Mcv: means corpuscular volume Alkphos\\
2.Alkaline phosphates\\
3.Sgpt: alamine aminotransferase\\
4.Sgot: aspartate aminotransferase\\
5.Gammagt: gamma-glut amyl Tranpeptidase\\
6.Drinks: number of half-pint equivalents of
alcoholic beverages Drunk per day\\
7.Selector: field used to split data into two sets. \\
The dataset is continuous, and there is no missing or
destroyed data.

\section{Fuzzy Inference Systems (FIS)}
\label{sec:fes}
\begin{figure}[h]
\centering
  \includegraphics[width=0.5\linewidth]{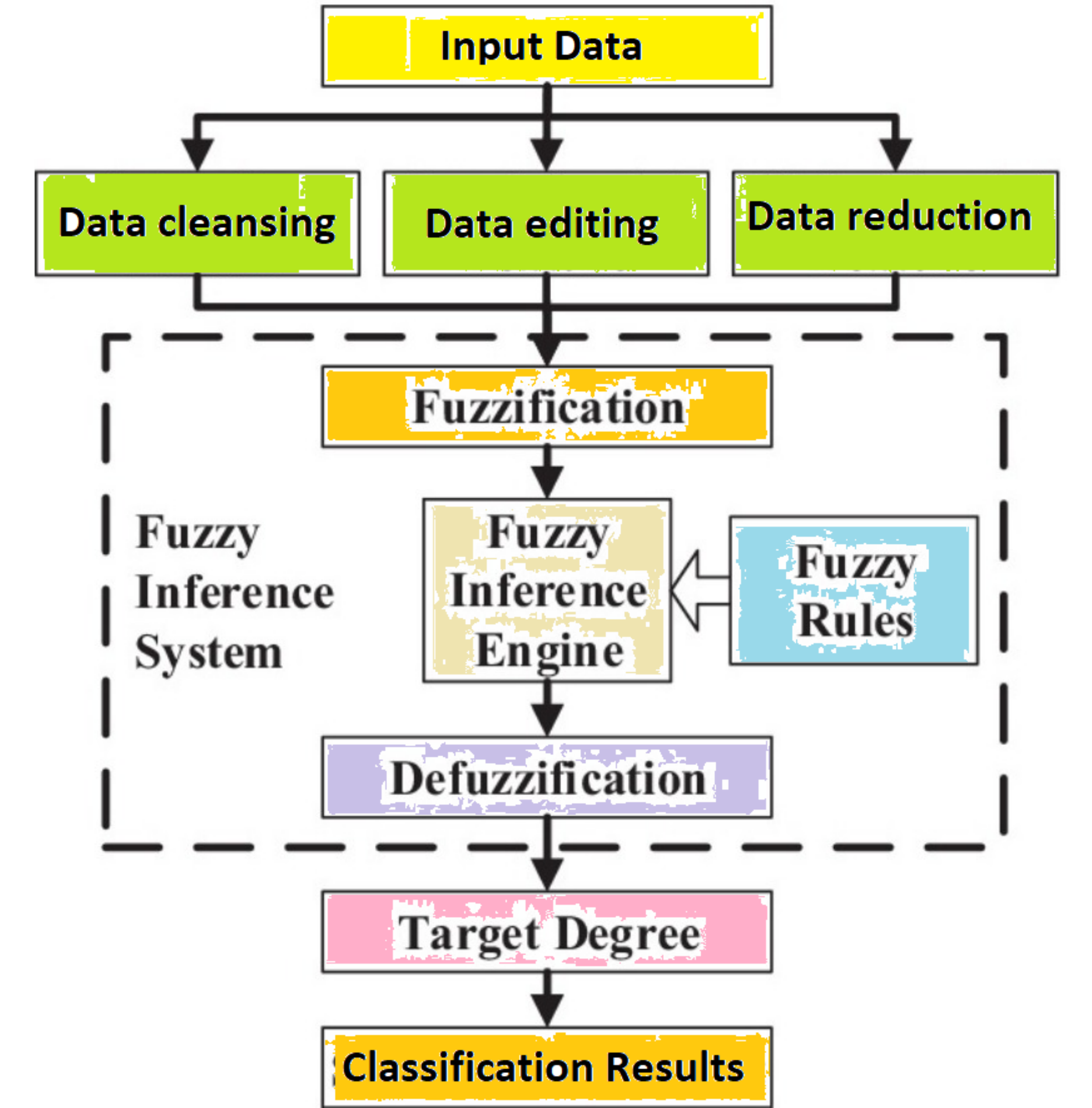}
  \caption{The scheme of the Liver disorders diagnotic Fuzzy Inference System}
  \label{fig:FIS}
\end{figure}

Initially, Zadeh \cite{zadeh1996soft} introduced the main theory of fuzzy logic as an approach for interpreting human knowledge that is not precise and well-defined. The process of fuzzification interface converts the crisp information into fuzzy linguistic values by various kinds of membership functions. The fuzzification can be regularly required in a fuzzy expert system considering the input values from surviving detectors are always deterministic numerical values. The inference generator demands fuzzy input and rules, and then it will produce fuzzy productions. Considerably, the fuzzy rule base should be in the figure of “IF-THEN” rules, including linguistic variables. The last part of a fuzzy expert system can be defuzzification which has the responsibility of performing crisp yield operations. The landscape of the fuzzy inference system can be represented in Figure \ref{fig:FIS}.

In the last three decades, Fuzzy rule-based systems is a subsidiary of Artificial intelligence fitted of interpreting complicated medical data. Their potential to employ significant relationship within a data set has been used in the diagnosis, treatment and predicting consequence in various clinical outlines. A survey of different artificial intelligence methods is exhibited in this part, along with the study of critical clinical applications of expert systems. The ability of artificial intelligence systems and has been explored in almost every field of medicine. Artificial neural network and knowledge based systems were the most regularly accepted analytical tool while additional AI systems such as evolutionary algorithms, swarm intelligence and hybrid systems have been handled in various clinical environments. It can be concluded that AI and expert systems have a high potential to be employed in almost all fields of medicine. Table \ref{table:medicine} shows the application of practical AI techniques such as fuzzy sets, neural networks, evolutionary algorithms, swarm intelligence for diagnosing a wide set of diseases. Table \ref{table:medicine} shows a short review of different kinds methods for diagnosing the Liver disorders in the last two decades.  
\begin{table}
\begin{center}
    \begin{tabular}{ p{3cm} | p{7cm} | p{3cm} | p{1cm} }
    \hline \hline
    \textbf{Authors} & \textbf{Methods} & \textbf{Disease} & \textbf{Year} \\ \hline
      Neshat et al. \cite{neshat2008designing,neshat2008fuzzy,neshat2010hopfield,neshat2014diagnosing,Neshat2013asurvey}& Bayesian parametric method and Parzen window non parametric method, Fuzzy Expert System, Hopfield Neural Network and Fuzzy Hopfield Neural Network & Liver Disease  &2008, 2009, 2010, 2013, 2014  \\ \hline
      Selvaraj et al.\cite{selvaraj2013improved}&particle swarm optimization  &Liver Disease &2013 \\ \hline
  Satarkar et al.\cite{satarkar2015fuzzy}&Fuzzy expert system &Liver Disease &2015 \\ \hline
    Hashemi et al. \cite{hashmi2015diagnosis} & fuzzy logic& Liver Disease &2015 \\ \hline
    Singh et al.\cite{singh2018efficient}&Principal Component Analysis and K-Nearest Neighbor (PCA-KNN) & Liver Disease &2018 \\ \hline
  Mirmozaffari et al. \cite{mirmozaffari2019developing}&expert system  & Liver Disease &2019 \\ \hline
   Kim et al. \cite{kim2014effective} & neural network and fuzzy neural network&Liver Cancer  & 2014 \\ \hline
  Das et al. \cite{das2018adaptive}&Adaptive fuzzy clustering-based texture analysis &Liver Cancer &2018 \\ \hline
   Xian et al. \cite{xian2010identification}& GLCM texture features and fuzzy SVM &Liver Tumors &2010 \\ \hline
    Polat et al. \cite{polat2007expert} & adaptive neuro-fuzzy inference system  & Diabetes Disease &2007  \\ \hline
    Polat et al. \cite{polat2006hepatitis}&artificial immune recognition system with fuzzy resource allocation  & Hepatitis Disease  & 2006 \\ \hline
    Chen et al.\cite{chen2011new}&local fisher discriminant analysis and support vector machines  & Hepatitis Disease &2011 \\ \hline
   Neshat et al. \cite{neshat2009designing, neshat2012hepatitis,neshat2009feshdd}& Adaptive Neural Network Fuzzy System, Hybrid Case Based Reasoning and PSO, Fuzzy expert system & Hepatitis B & 2009, 2012 \\ \hline
   Adeli et al. \cite{adeli2013new}&Genetic algorithm and adaptive network fuzzy inference system &Hepatitis  &2013 \\ \hline
  Ahmad et al. \cite{ahmad2018intelligent,ahmad2019automated}&adaptive neuro-fuzzy inference system, Multilayer Mamdani Fuzzy Inference System &Hepatitis Disease & 2018, 2019 \\ \hline
    \end{tabular}
    \caption{A briefly survey of the AI method applications for diagnosing the Liver disorders.}
\label{table:medicine}
\end{center}
\end{table}

\section{Adaptive Neural Fuzzy Inference System (ANFIS)}

\begin{figure}[h]
\centering
  \includegraphics[width=0.6\linewidth]{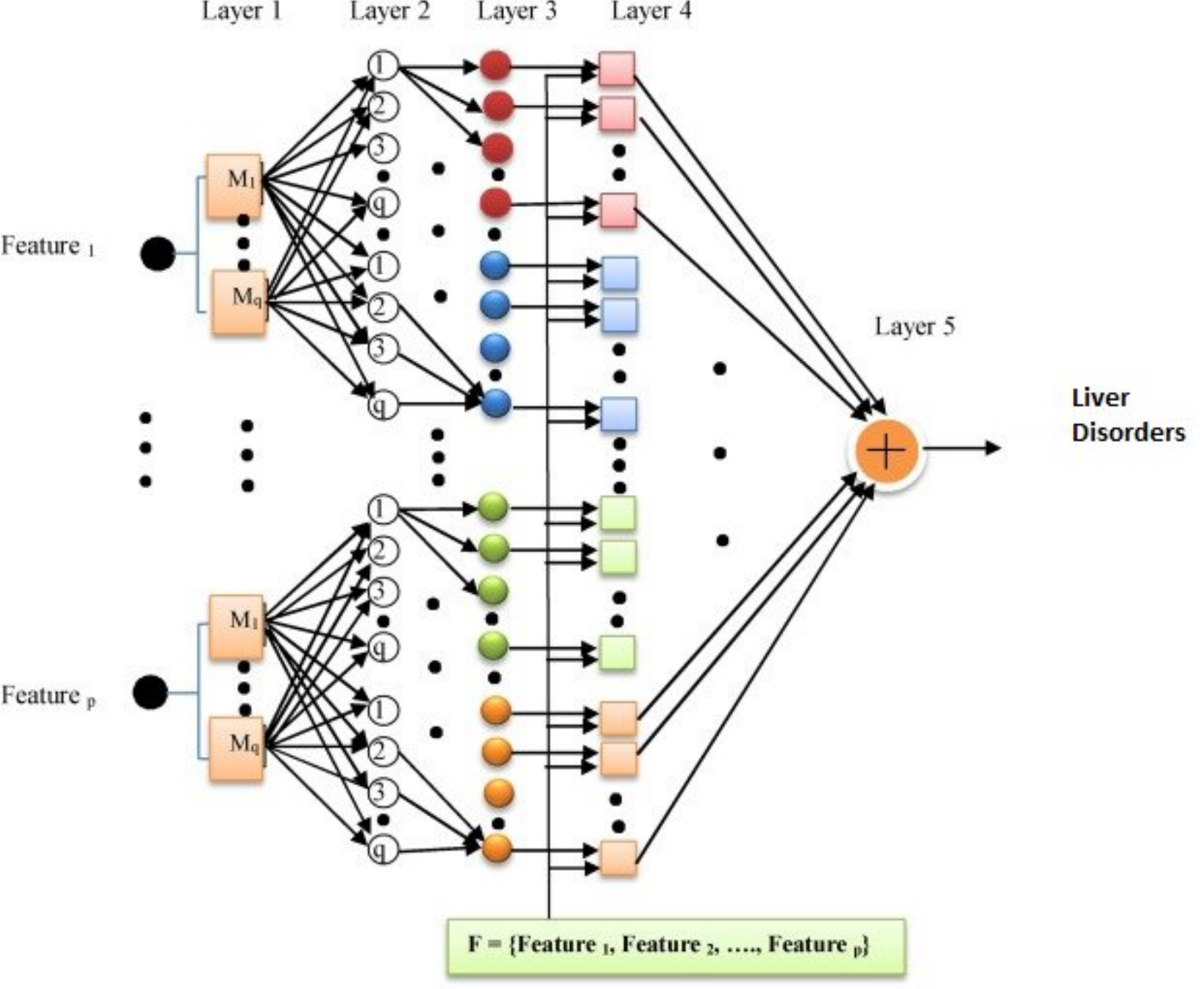}
  \caption{The five layers architecture  of an Adaptive Neuro-Fuzzy Inference System from \cite{nilashi2019predictive}}
  \label{fig:ANFIS}
\end{figure}

An adaptive neuro-fuzzy inference system (ANFIS) can be a class of artificial neural network (ANN) that is worked in regard to Takagi–Sugeno fuzzy inference system. The system was developed at the beginning of the 1990s \cite{jang1991fuzzy}. Since it combines both ANN and fuzzy logic principles, it holds the potential to catch the advantages of both in a unique framework. Its fuzzy inference system (FIS) corresponds to a collection of fuzzy rules (IF–THEN) which have learning inclination to approximate nonlinear functions. Consequently, ANFIS is supposed to be a general estimator. For practising the ANFIS more efficiently and optimally, one can handle the most useful parameters taken by genetic algorithm\cite{tahmasebi2012hybrid}. 
It is conceivable to distinguish two parts in the network structure, namely basis and consequence parts. In more details, the architecture is comprised of five layers. The first layer receives the input values and determines the membership functions referring to them. It is generally called fuzzification layer. The membership degrees of each function are calculated by applying the premise parameter set, namely {a,b,c}. The second layer is responsible for making the firing strengths for the rules. Due to its responsibility, the second layer is expressed as "rule layer". The role of the third layer is to normalize the measured firing strengths by diving each value for the total firing strength. The fourth layer practices as input the normalized values and the result parameter set {p,q,r}. The values yielded by this layer are the defuzzificated ones, and also those values are transferred to the last layer to replace the final output \cite{karaboga2018adaptive}. Figure \ref{fig:ANFIS} shows a deep landscape of ANFIS architecture. Table \ref{table:medicine} shows the previous applied methods for diagnosing the liver disorders.

\section{Canonical Particle Swarm Optimization}
Particle swarm optimisation (PSO) was proposed by Kennedy and Eberhart \cite{trelea2003particle}, driven by the operation of social animals relationships in groups, such as bird and fish schooling or ant colonies. This metaheuristic idea accompanies in the interaction among members to share their achieved knowledge. PSO has been performed in various regions in optimisation and mixture with other existing algorithms. This method obtains the exploration of the optimal solution through particles, whose trajectories are mitigated by a stochastic and a deterministic component. Each solution called particle is influenced by its ‘best’ gave position and the population ‘best’ situation, but commands to walking randomly. A particle $i$ is distinguished by its status vector,$ x_i$, and its velocity vector,$ v_i$. Every iteration, each particle coordinates its situation based on the new velocity as:

\begin{figure}[h]
\centering
  \includegraphics[width=0.6\linewidth]{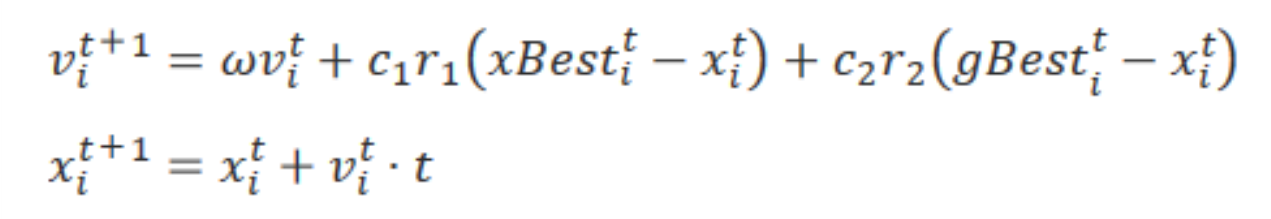}
  \end{figure}
where $xBest$ and $gBest$ indicate the best particle location and best group situation and the parameters $w, c_1, c_2, r_1$ and $r_2$ are respectively inertia weight, two positive constants and two random parameters within [0, 1]. In the baseline PSO, $w$ is chosen as a unit, but an enhancement of the PSO is observed in its inertial implementation using $w \in [0.5 0.9]$. Regularly, maximum and minimum velocity values are also described, and originally, the particles are assigned randomly to boost the search in all tolerable locations. Despite all the advantages of PSO like fast convergence and powerful in global search, its control parameters need to be tuned during the global search. There are plenty of proposed ideas to adjust the control parameters which have been called adaptive PSO \cite{zhan2009adaptive,neshat2010aipso,neshat2012new,hu2012adaptive,neshat2013faipso,lin2019adaptive}. Moreover, various meta-heuristic methods are combined with PSO to create a successful hybrid search technique \cite{naka2003hybrid,zahara2009hybrid,wang2018hybrid}. One of the benefits of PSO over other derivative-free approaches is the diminished the number of parameters to adjust and restrictions acceptance.    

\section{Adaptive Neural PSO and FIS}

\subsection{Adaptive 1+1EA}
Shortly after introducing the genuine random search as a stochastic optimization algorithm, it was identified those adaptive algorithms where the sampling distribution is adapted (as encountered to pure random search) throughout the course of the optimization can be essential for the efficient optimization process. One of the most used adaptive search techniques adjusts the step-size utilizing the subsequent idea:
the step-size is raised after a successful step (enhancing exploration) and decreased after a failure(developing exploitation) to keep a success probability of roughly $1/5$, grow the step-size if the success probability is more extensive than $1/5$ and decreases it oppositely.The original 1+1EA with adaptive mutation step size can be seen in the Algorithm 1 \cite{DBLP:conf/gecco/Auger09}.

\begin{figure}[h]
\centering

  \includegraphics[width=0.6\linewidth]{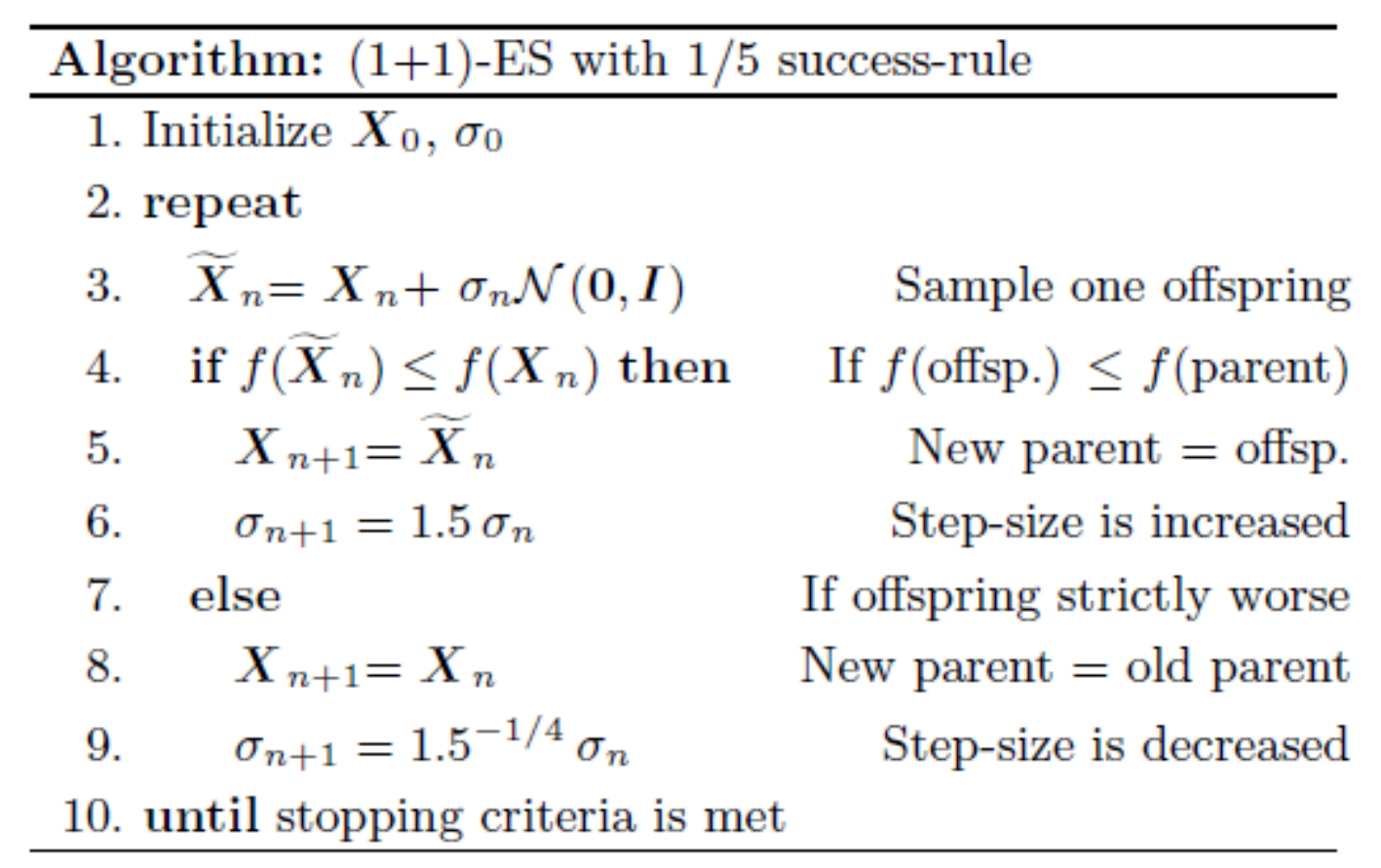}
    
\end{figure}
In the second version of the adaptive 1+1EA (Algorithm variant 1), it can be seen that we keep the mutation step size if there is not any improvement after applying mutation.   
\begin{figure}[h]
\centering
  \includegraphics[width=0.6\linewidth]{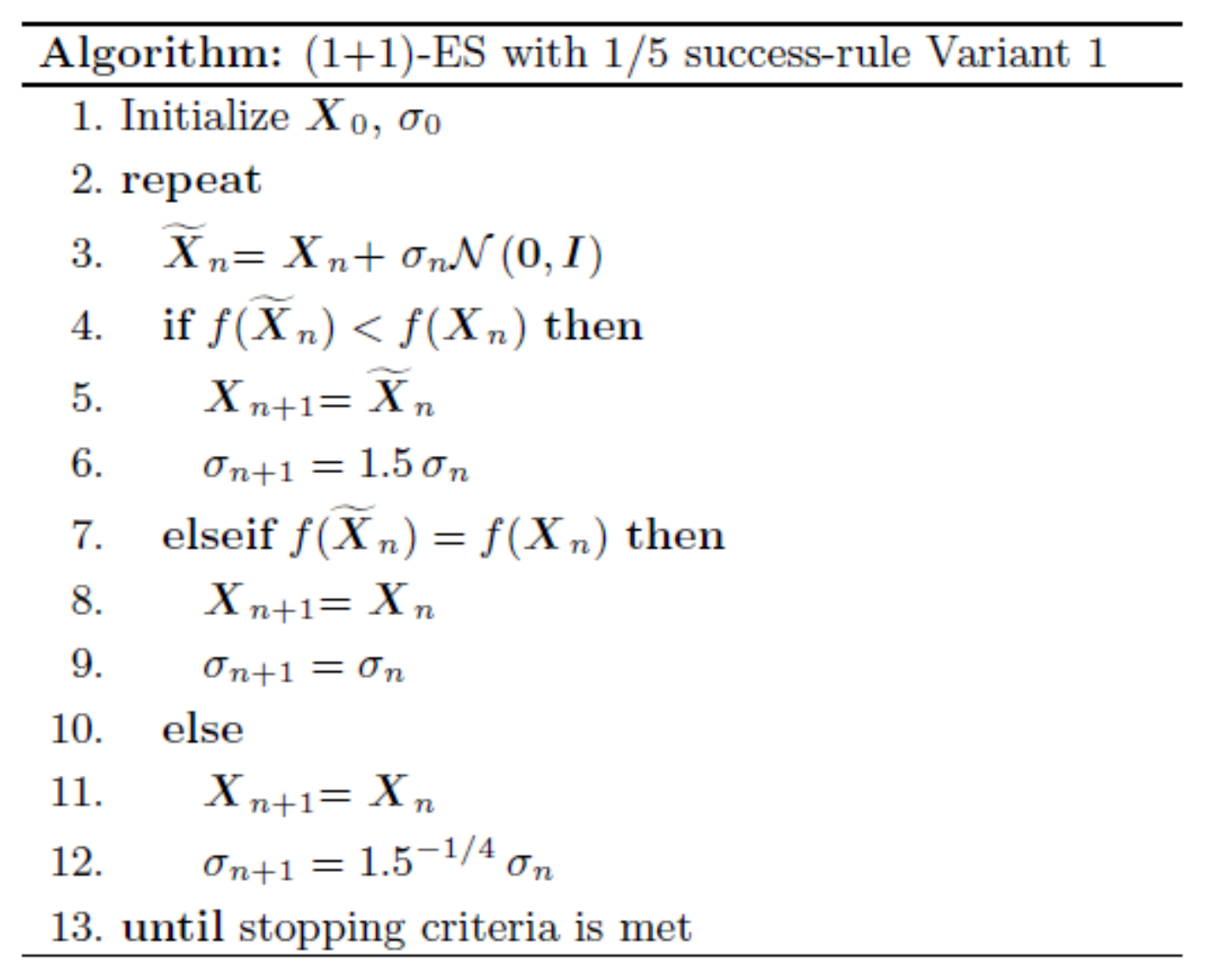}
    \label{fig:Algorithm2}
\end{figure}
We employ both version of adaptive 1+1EA for tuning the control parameters of PSO including $c_1$, $c_2$ and $w$ by 100 generations. The mutation probability is $1/N$ where $N$ is the length of decision variables. 

\subsection{Adaptive Neuro PSO plus FIS}
In this article, we propose a new adaptive particle swarm optimization which is combined with an inference system for diagnosing liver disorders. As the original PSO has some drawbacks like premature convergence and disables to have a robust global search, we apply a fast and effective evolutionary algorithm that is equipped by an adaptive mechanism called 1+1EA (with 1/5). This method is run every other ten iterations of PSO run to adjust the control parameters. In the next layer of the hybrid method, PSO which is a population-based method that can be an effective swarm intelligence method is employed to tune the features of the inference system like the number of fuzzy rules$[1-10]$, the number of fuzzy membership function $[1-5]$ in each input variable and also the type of fuzzy membership functions (triangle, gaussian and trapezoidal) and the range of each membership. 
In predominating, experts in a particular field is able to make the rules and membership function because the definition of these is generally affected by individual decisions. While fuzzy rules indicate approximately straightforward to obtain by them, the MFs imply doubting to complete. Besides, attuning of MFs can be a time-consuming process.  These characters perform evolutionary algorithms such as Genetic Algorithms (GAs), particle swarm optimization (PSO), better choices for searching these spaces \cite{muthukaruppan2012hybrid}.  Finally, a multi-layer feed-forward neural network is used to create the inference system based on the knowledge-based dataset. 

\begin{figure}[h]
\centering
  \includegraphics[width=\linewidth]{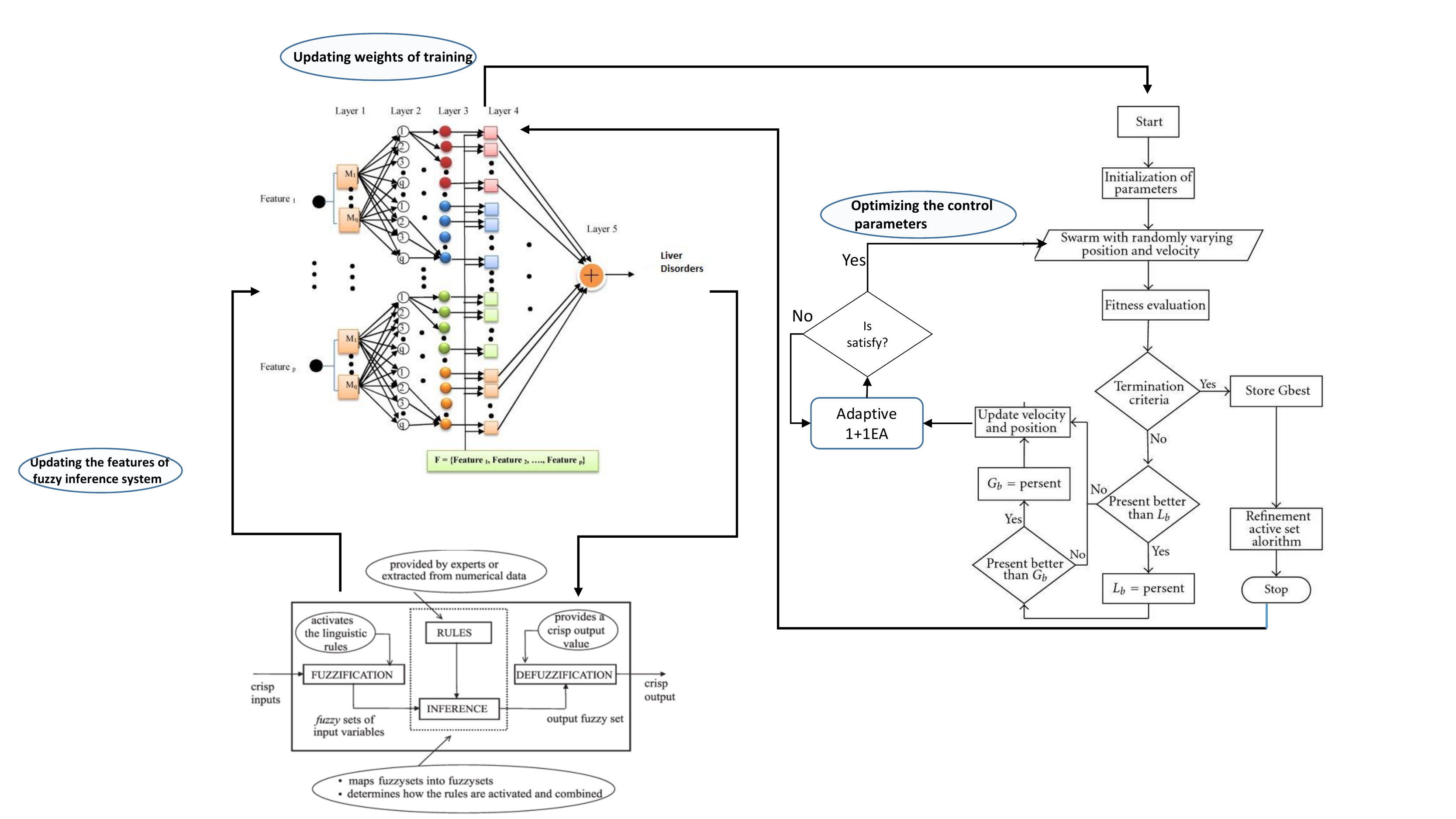}
  \caption{The architecture of an Adaptive Neuro PSO combined with a fuzzy inference system}
  \label{fig:ANPSO}
\end{figure}

\section{Experimental outcomes}
\begin{figure}[t]
\subfloat[]{
\includegraphics[clip,width=0.49\columnwidth]{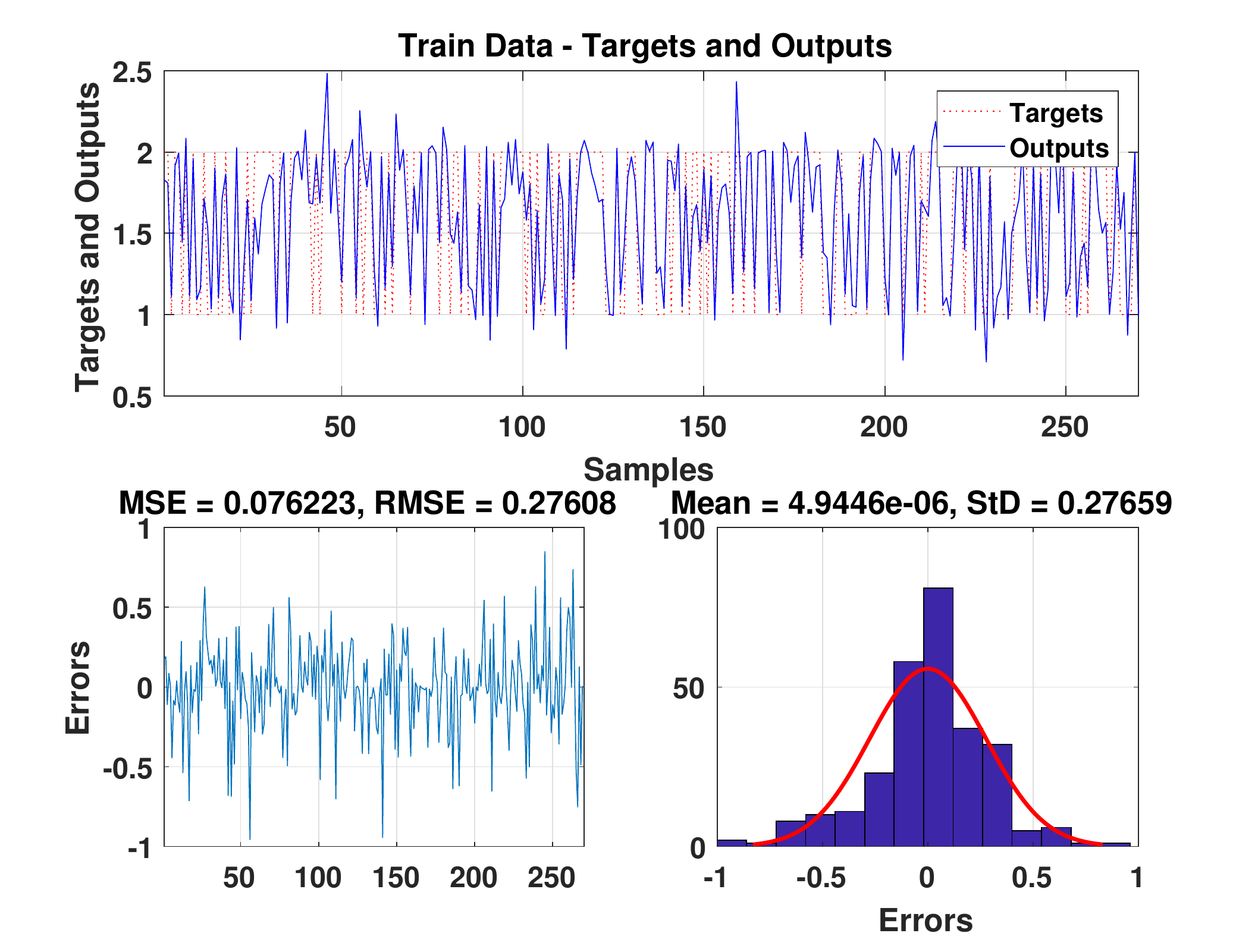}}
\subfloat[]{
\includegraphics[clip,width=0.49\columnwidth]{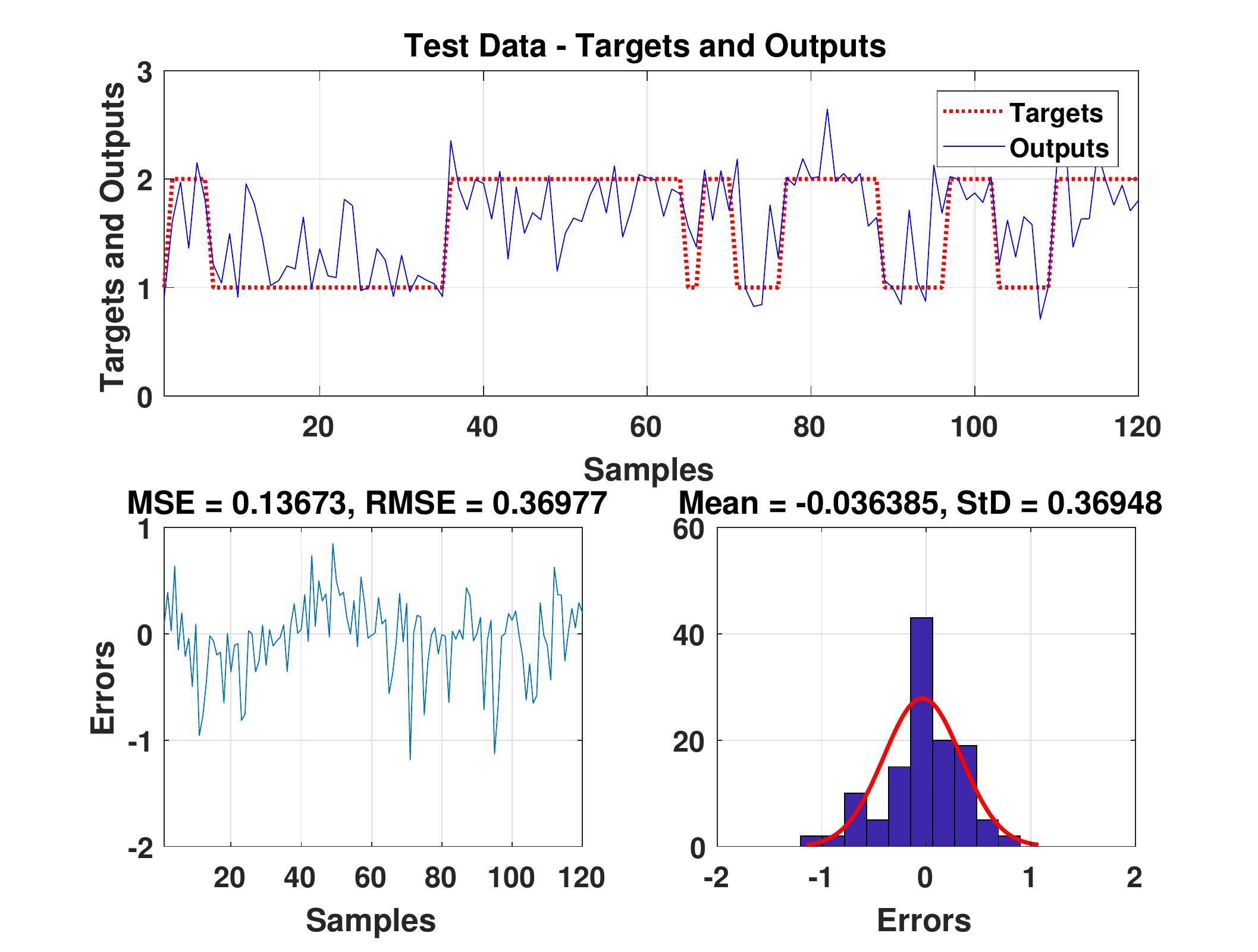}}
\caption{The training and testing results and error of an original ANFIS performance for diagnosing Liver disorders.}%
\label{fig:ANFIS_TEST_TRAIN}%
\end{figure}

The original ANFIS and some hybrids of optimization methods performances are assessed by the dataset of liver disorders. The given evaluation criteria can be the mean square error (MSE) and MSE root  (RMSE) of the targets and predicted outputs. Figure \ref{fig:ANFIS_TEST_TRAIN} explains the MSE of both trained and tested data of the original ANFIS performance for one run as a tangible example. The error distribution can be in regard to a normal distribution with an approximately wide variance which shows that the applied ANFIS need to be modified. The average RMSE of ANFIS train and test are specified at $0.27$ and $0.37$. Meanwhile, two different optimization methods of ANFIS are compared. The results show that the hybrid optimization method performance is better than the back-propagation approach. 

Table \ref{table:optimizationmethod} shows the comparison of the achieved results of both ANFIS optimization method. Furthermore, Table \ref{table:all-results} represents the RMSE of applied hybrid methods with original ANFIS. We can see that the best performances are related to adaptive PSO and original PSO with ANFIS.
The average RMSE of ANPSO-ANFIS training and testing are at $0.042$ and $0.057$. These results show an acceptable development for the newly proposed method. 

\begin{figure}[h]
\centering
  \includegraphics[width=0.6\linewidth]{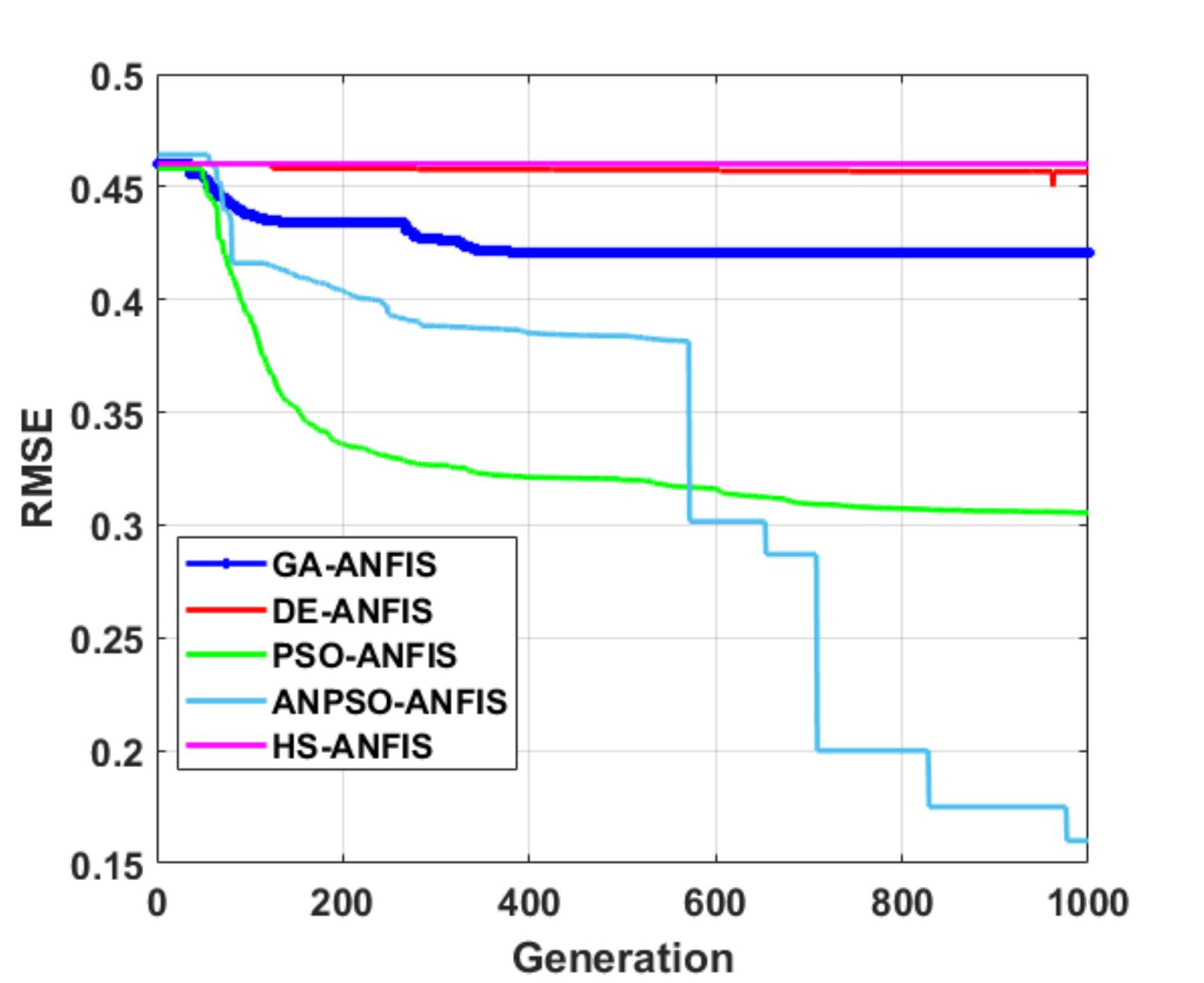}
  \caption{The performance of applied PSO to tune the FIS parameters and reduce the RMSE.}%
  \label{fig:ANFIS_PSO_per}
  \end{figure}

In the meantime, Figure \ref{fig:ANFIS_PSO_per} shows how adaptive PSO is able to tune the different parameters of the used inference system. The overall process of hybrid adaptive PSO and other applied methods can be shown in Figure \ref{fig:ANFIS_PSO_per}. In spite of an initial fast convergence of PSO-ANFIS, the ANPSO-ANFIS can overcome all methods considerably.






\begin{table}[]
\centering
\begin{tabular}{|l|l|l|l|l|l|l|l|l|}
\hline
\multicolumn{9}{ |c| }{\textbf{ANFIS}}\\  \hline
\multicolumn{5}{ |c| }{\textbf{RMSE}}& \multicolumn{4}{ |c| }{\textbf{R-value}}\\  \hline
        
        &  \multicolumn{2}{ |c| }{Hybrid} &  \multicolumn{2}{| c| }{back-propagation}     & \multicolumn{2}{ |c| }{Hybrid}  &        \multicolumn{2}{ |c| }{back-propagation}       \\ \hline
        & Train  & Test   & Train           & Test & Train   & Test  & Train           & Test  \\ \hline
max     & 0.29   & 0.481  & 0.38            & 0.42 & 0.85    & 0.96  & 0.68            & 0.68  \\ \hline
min     & 0.262  & 0.33   & 0.36            & 0.4  & 0.75    & 0.57  & 0.63            & 0.54  \\ \hline
average & 0.272  & 0.3732 & 0.37            & 0.41 & 0.816   & 0.706 & 0.654           & 0.582 \\ \hline
\end{tabular}
\caption{The statistical results of 30 independent runs for an original ANFIS with two different optimization methods. }
\label{table:optimizationmethod}
\end{table}
\begin{table}[]
 \centering
\scalebox{0.9}{
\begin{tabular}{|l|l|l|l|l|l|l|l|l|l|l|l|l|}
\hline
&  \multicolumn{2}{ |c }{ANFIS}        &  \multicolumn{2}{ |c }{PSO-ANFIS}         & \multicolumn{2}{ |c }{DE-ANFIS}        &\multicolumn{2}{ |c }{GA-ANFIS}        &\multicolumn{2}{ |c }{HS-ANFIS }        & \multicolumn{2}{ |c| }{ANPSO-ANFIS}              \\ \hline \hline
& Train & Test & Train & Test & Train & Test & Train & Test & Train & Test & Train & Test   \\ \hline
max     & 0.29  & 0.481     & 0.24  & 0.28     & 0.38  & 0.48     & 0.31  & 0.39     & 0.38  & 0.41        & 0.05  & 0.08   \\ \hline
min     & 0.262 & 0.33      & 0.19  & 0.27     & 0.25  & 0.32     & 0.26  & 0.31     & 0.29  & 0.31        & 0.006 & 0.0032 \\ \hline
average & 0.272 & 0.3732    & 0.226 & 0.277    & 0.293 & 0.373    & 0.291 & 0.359    & 0.355 & 0.366       & 0.042 & 0.057\\  \hline
\end{tabular}
}
\caption{The statistical results based on RMSE of 30 independent runs for an original ANFIS compared with other hybrid ideas. }
\label{table:all-results}
\end{table}
\begin{figure}[t]
\subfloat[]{
\includegraphics[clip,width=0.49\columnwidth]{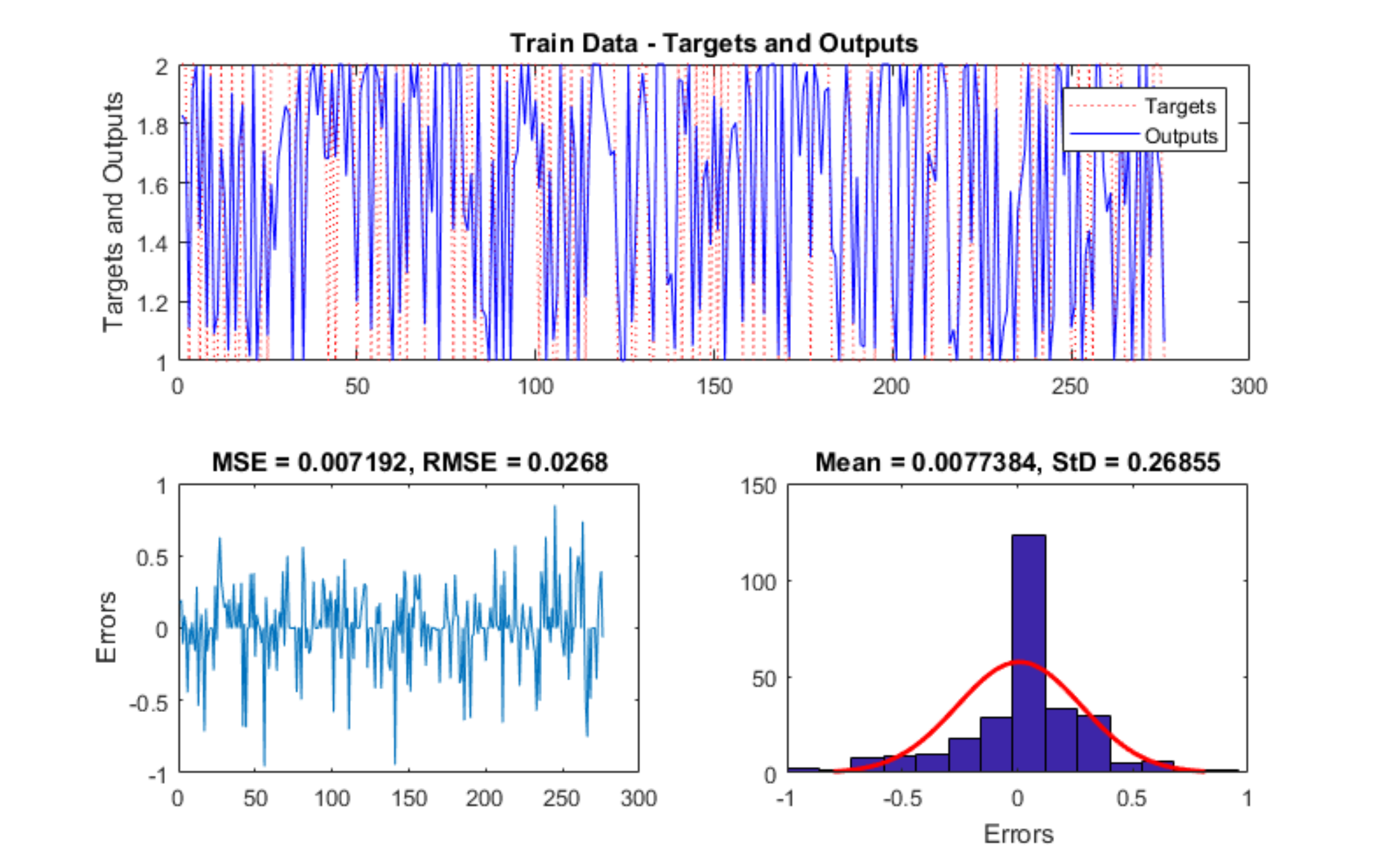}}
\subfloat[]{
\includegraphics[clip,width=0.49\columnwidth]{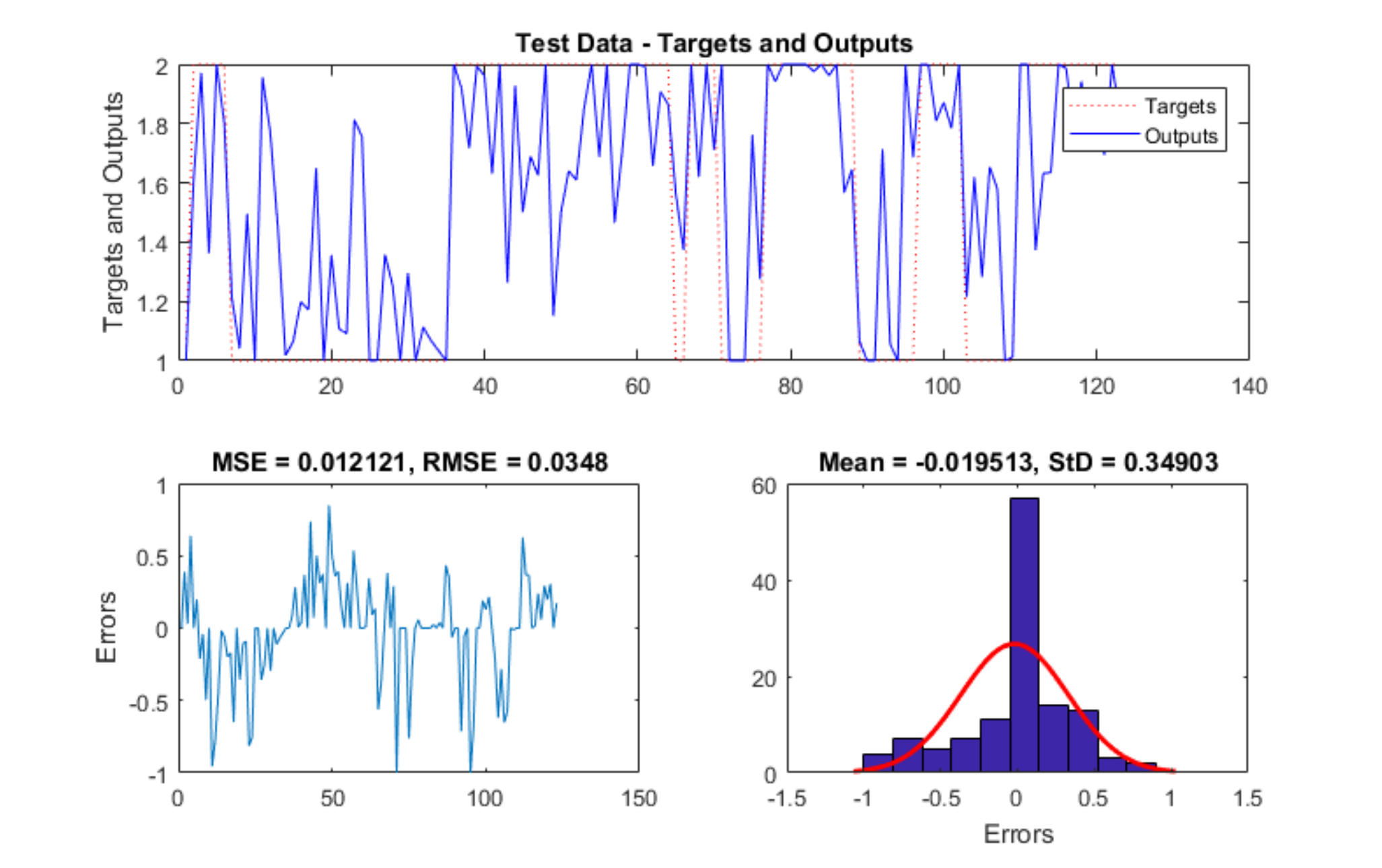}}
\caption{The training and testing results and error of the proposed ANPSO-ANFIS performance for diagnosing Liver disorders.}%
\label{fig:ANFIS_PSO_TEST_TRAIN}%
\end{figure}

\section{  Conclusions}
\label{sec:con}
In this article, a hybrid adaptive neural PSO is proposed and implemented in Matlab’s Simulink because of recognising the liver disorders. According to the practical results, the performance of the proposed method is increased by $20\%$ compared with the original ANFIS and other hybrid optimization techniques based on the dataset can be performed. The expansion of the important characteristics and fuzzy rules are taken by applying the statistical analysis. The significance of recognising meaningful and appropriate fuzzy rules without the support of the professionals exhibits the possibility of knowledge discovery. The main advantages of the FIS as a knowledge acquisition tool are the following: firstly, the adaptive number of rules are evaluated. Secondly, the obtained rules can be efficiently represented. These results propose encouraging research areas employing adaptive PSO and fuzzy inference system in different classification problems. The hybrid proposed system is able to overcome beforehand considered methods in terms of both precision and portability.

\bibliographystyle{unsrt}  
\bibliography{references}

\begin{thebibliography}{10}

\bibitem{feigenbaum1981expert}
Edward~A Feigenbaum.
\newblock Expert systems in the 1980s.
\newblock {\em State of the art report on machine intelligence. Maidenhead:
  Pergamon-Infotech}, 1981.

\bibitem{polat2006diagnosis}
Kemal Polat, Salih G{\"u}ne{\c{s}}, and S{\"u}layman Tosun.
\newblock Diagnosis of heart disease using artificial immune recognition system
  and fuzzy weighted pre-processing.
\newblock {\em Pattern Recognition}, 39(11):2186--2193, 2006.

\bibitem{csahan2007new}
Seral {\c{S}}ahan, Kemal Polat, Halife Kodaz, and Salih G{\"u}ne{\c{s}}.
\newblock A new hybrid method based on fuzzy-artificial immune system and k-nn
  algorithm for breast cancer diagnosis.
\newblock {\em Computers in Biology and Medicine}, 37(3):415--423, 2007.

\bibitem{adeli2010fuzzy}
Ali Adeli and Mehdi Neshat.
\newblock A fuzzy expert system for heart disease diagnosis.
\newblock In {\em Proceedings of International Multi Conference of Engineers
  and Computer Scientists, Hong Kong}, volume~1, pages 28--30, 2010.

\bibitem{neshat2015new}
Mehdi Neshat, Ghodrat Sepidname, Amin Eizi, and Amanollah Amani.
\newblock A new skin color detection approach based on fuzzy expert system.
\newblock {\em Indian Journal of Science and Technology}, 8:1--11, 2015.

\bibitem{moya2019fuzzy}
Alejandro Moya, Elena Navarro, Javier Ja{\'e}n, and Pascual Gonz{\'a}lez.
\newblock Fuzzy-description logic for supporting the rehabilitation of the
  elderly.
\newblock {\em Expert Systems}, page e12464, 2019.

\bibitem{neshat2011designing}
Mehdi Neshat and Ali Adeli.
\newblock Designing a fuzzy expert system to predict the concrete mix design.
\newblock In {\em 2011 IEEE International Conference on Computational
  Intelligence for Measurement Systems and Applications (CIMSA) Proceedings},
  pages 1--6. IEEE, 2011.

\bibitem{neshat2012predication}
Mehdi Neshat, Ali Adeli, Ghodrat Sepidnam, and Mehdi Sargolzaei.
\newblock Predication of concrete mix design using adaptive neural fuzzy
  inference systems and fuzzy inference systems.
\newblock {\em The International Journal of Advanced Manufacturing Technology},
  63(1-4):373--390, 2012.

\bibitem{yuan2014prediction}
Zhe Yuan, Lin-Na Wang, and Xu~Ji.
\newblock Prediction of concrete compressive strength: Research on hybrid
  models genetic based algorithms and anfis.
\newblock {\em Advances in Engineering Software}, 67:156--163, 2014.

\bibitem{chiew2017fuzzy}
Fei~Ha Chiew, Chee~Khoon Ng, Kok~Chin Chai, and Kai~Meng Tay.
\newblock A fuzzy adaptive resonance theory-based model for mix proportion
  estimation of high-performance concrete.
\newblock {\em Computer-Aided Civil and Infrastructure Engineering},
  32(9):772--786, 2017.

\bibitem{pourahmad2012service}
Ali~Akbar Pourahmad, Mehdi Neshat, and Ahmad Baghi.
\newblock Service quality assessment in the academic library: Use of hybrid
  fuzzy expert system.
\newblock {\em African Journal of Business Management}, 6(46):11511--11529,
  2012.

\bibitem{pourahmad2016using}
Ali~Akbar Pourahmad, Mehdi Neshat, and Mohammad~Reza Hasani.
\newblock Using libqual model for improving the level of students’
  satisfaction from quality of services in academic libraries: A case study in
  north khorasan province, iran.
\newblock {\em Journal of Information \& Knowledge Management}, 15(01):1650011,
  2016.

\bibitem{raeesi2018quality}
Saeed Raeesi.
\newblock Quality assessment of ilam university of medical sciences website
  from the users’ viewpoints according to webqual model.
\newblock {\em scientific journal of ilam university of medical sciences},
  26(4):53--63, 2018.

\bibitem{du2018fuzzy}
Engelina Du~Plessis, Juan~Carlos Martin, Concepcion Roman, and Elmarie
  Slabbert.
\newblock Fuzzy logic to assess service quality at arts festivals.
\newblock {\em Event Management}, 22(4):501--516, 2018.

\bibitem{shafii2016assessment}
Milad Shafii, Sima Rafiei, Fatemeh Abooee, Mohammad~Amin Bahrami, Mojtaba
  Nouhi, Farhad Lotfi, and Khatere Khanjankhani.
\newblock Assessment of service quality in teaching hospitals of yazd
  university of medical sciences: Using multi-criteria decision making
  techniques.
\newblock {\em Osong public health and research perspectives}, 7(4):239--247,
  2016.

\bibitem{neshat2011fhesmm}
Mehdi Neshat, Ahmad Baghi, Ali~Akbar Pourahmad, Ghodrat Sepidnam, Mehdi
  Sargolzaei, and Azra Masoumi.
\newblock Fhesmm: Fuzzy hybrid expert system for marketing mix model.
\newblock {\em International Journal of Computer Science Issues(IJCSI)}, 8(6),
  2011.

\bibitem{neshat2016designing}
Mehdi Neshat, Ali~Akbar Pourahmad, and Mohammad~Reza Hasani.
\newblock Designing an adaptive neuro fuzzy inference system for prediction of
  customers satisfaction.
\newblock {\em Journal of Information \& Knowledge Management}, 15(04):1650037,
  2016.

\bibitem{rudzewicz2015quality}
Adam Rudzewicz.
\newblock Quality of banking services from the perspective of the polish
  customers.
\newblock {\em Zeszyty Naukowe Wy{\.z}szej Szko{\l}y Ekonomiczno-Spo{\l}ecznej
  w Ostro{\l}{\k{e}}ce}, 1(19):65--73, 2015.

\bibitem{deveci2018interval}
Muhammet Deveci, Ender {\"O}zcan, Robert John, and Sultan~Ceren {\"O}ner.
\newblock Interval type-2 hesitant fuzzy set method for improving the service
  quality of domestic airlines in turkey.
\newblock {\em Journal of Air Transport Management}, 69:83--98, 2018.

\bibitem{shi1998modified}
Yuhui Shi and Russell Eberhart.
\newblock A modified particle swarm optimizer.
\newblock In {\em 1998 IEEE international conference on evolutionary
  computation proceedings. IEEE world congress on computational intelligence
  (Cat. No. 98TH8360)}, pages 69--73. IEEE, 1998.

\bibitem{ratnaweera2004self}
Asanga Ratnaweera, Saman~K Halgamuge, and Harry~C Watson.
\newblock Self-organizing hierarchical particle swarm optimizer with
  time-varying acceleration coefficients.
\newblock {\em IEEE Transactions on evolutionary computation}, 8(3):240--255,
  2004.

\bibitem{angeline1998using}
Peter~J Angeline.
\newblock Using selection to improve particle swarm optimization.
\newblock In {\em 1998 IEEE International Conference on Evolutionary
  Computation Proceedings. IEEE World Congress on Computational Intelligence
  (Cat. No. 98TH8360)}, pages 84--89. IEEE, 1998.

\bibitem{brits2002niching}
Riaan Brits, Andries~P Engelbrecht, and F~Van~den Bergh.
\newblock A niching particle swarm optimizer.
\newblock In {\em Proceedings of the 4th Asia-Pacific conference on simulated
  evolution and learning}, volume~2, pages 692--696. Singapore: Orchid Country
  Club, 2002.

\bibitem{farokhzad2016novel}
Maryam~Rezaei Farokhzad and Laya Ebrahimi.
\newblock A novel adaptive neuro fuzzy inference system for the diagnosis of
  liver disease.
\newblock {\em International Journal of Academic Research in Computer
  Engineering}, 1(1):61--66, 2016.

\bibitem{zadeh1996soft}
Lotfi~A Zadeh.
\newblock Soft computing and fuzzy logic.
\newblock In {\em Fuzzy Sets, Fuzzy Logic, and Fuzzy Systems: Selected Papers
  by Lotfi a Zadeh}, pages 796--804. World Scientific, 1996.

\bibitem{neshat2008designing}
Mehdi Neshat, Mehdi Yaghobi, and Mohammad Naghibi.
\newblock Designing an expert system of liver disorders by using neural network
  and comparing it with parametric and nonparametric system.
\newblock In {\em 2008 5th International Multi-Conference on Systems, Signals
  and Devices}, pages 1--6. IEEE, 2008.

\bibitem{neshat2008fuzzy}
M~Neshat, M~Yaghobi, MB~Naghibi, and A~Esmaelzadeh.
\newblock Fuzzy expert system design for diagnosis of liver disorders.
\newblock In {\em 2008 International Symposium on Knowledge Acquisition and
  Modeling}, pages 252--256. IEEE, 2008.

\bibitem{neshat2010hopfield}
Mehdi Neshat and Abas~E Zadeh.
\newblock Hopfield neural network and fuzzy hopfield neural network for
  diagnosis of liver disorders.
\newblock In {\em 2010 5th IEEE International Conference Intelligent Systems},
  pages 162--167. IEEE, 2010.

\bibitem{neshat2014diagnosing}
Mehdi Neshat, Azra Masoumi, Mina Rajabi, and Hassan Jafari.
\newblock Diagnosing hepatitis disease by using fuzzy hopfield neural network.
\newblock {\em Annual Research \& Review in Biology}, pages 2709--2721, 2014.

\bibitem{Neshat2013asurvey}
Mehdi Neshat, Ali Adeli, and Azra Masoumi.
\newblock A survey on artificial intelligence and expert system for liver
  disorders.
\newblock {\em ARPN Journal of Systems and Software}, 3(2), 2013.

\bibitem{selvaraj2013improved}
Gunasundari Selvaraj and S~Janakiraman.
\newblock Improved feature selection based on particle swarm optimization for
  liver disease diagnosis.
\newblock In {\em International Conference on Swarm, Evolutionary, and Memetic
  Computing}, pages 214--225. Springer, 2013.

\bibitem{satarkar2015fuzzy}
SL~Satarkar and MS~Ali.
\newblock Fuzzy expert system for the diagnosis of common liver disease.
\newblock {\em International Engineering Journal For Research \& Development},
  1(1):2--7, 2015.

\bibitem{hashmi2015diagnosis}
Asma Hashmi and Muhammad~Saleem Khan.
\newblock Diagnosis blood test for liver disease using fuzzy logic.
\newblock {\em International Journal of Sciences: Basic and Applied Research
  (IJSBAR)}, 20(1):151--183, 2015.

\bibitem{singh2018efficient}
Aman Singh and Babita Pandey.
\newblock An efficient diagnosis system for detection of liver disease using a
  novel integrated method based on principal component analysis and k-nearest
  neighbor (pca-knn).
\newblock In {\em Intelligent Systems: Concepts, Methodologies, Tools, and
  Applications}, pages 1015--1030. IGI Global, 2018.

\bibitem{mirmozaffari2019developing}
Mirpouya Mirmozaffari.
\newblock Developing an expert system for diagnosing liver diseases.
\newblock {\em European Journal of Engineering Research and Science},
  4(3):1--5, 2019.

\bibitem{kim2014effective}
Sangman Kim, Seungpyo Jung, Youngju Park, Jihoon Lee, and Jusung Park.
\newblock Effective liver cancer diagnosis method based on machine learning
  algorithm.
\newblock In {\em 2014 7th International Conference on Biomedical Engineering
  and Informatics}, pages 714--718. IEEE, 2014.

\bibitem{das2018adaptive}
Amita Das, Priti Das, Soumya~S Panda, and Sukanta Sabut.
\newblock Adaptive fuzzy clustering-based texture analysis for classifying
  liver cancer in abdominal ct images.
\newblock {\em International Journal of Computational Biology and Drug Design},
  11(3):192--208, 2018.

\bibitem{xian2010identification}
Guang-ming Xian.
\newblock An identification method of malignant and benign liver tumors from
  ultrasonography based on glcm texture features and fuzzy svm.
\newblock {\em Expert Systems with Applications}, 37(10):6737--6741, 2010.

\bibitem{polat2007expert}
Kemal Polat and Salih G{\"u}ne{\c{s}}.
\newblock An expert system approach based on principal component analysis and
  adaptive neuro-fuzzy inference system to diagnosis of diabetes disease.
\newblock {\em Digital Signal Processing}, 17(4):702--710, 2007.

\bibitem{polat2006hepatitis}
Kemal Polat and Salih G{\"u}ne{\c{s}}.
\newblock Hepatitis disease diagnosis using a new hybrid system based on
  feature selection (fs) and artificial immune recognition system with fuzzy
  resource allocation.
\newblock {\em Digital Signal Processing}, 16(6):889--901, 2006.

\bibitem{chen2011new}
Hui-Ling Chen, Da-You Liu, Bo~Yang, Jie Liu, and Gang Wang.
\newblock A new hybrid method based on local fisher discriminant analysis and
  support vector machines for hepatitis disease diagnosis.
\newblock {\em Expert Systems with Applications}, 38(9):11796--11803, 2011.

\bibitem{neshat2009designing}
Mehdi Neshat and Mehdi Yaghobi.
\newblock Designing a fuzzy expert system of diagnosing the hepatitis b
  intensity rate and comparing it with adaptive neural network fuzzy system.
\newblock In {\em Proceedings of the World Congress on Engineering and Computer
  Science}, volume~2, pages 797--802, 2009.

\bibitem{neshat2012hepatitis}
Mehdi Neshat, Mehdi Sargolzaei, Adel Nadjaran~Toosi, and Azra Masoumi.
\newblock Hepatitis disease diagnosis using hybrid case based reasoning and
  particle swarm optimization.
\newblock {\em ISRN Artificial Intelligence}, 2012, 2012.

\bibitem{neshat2009feshdd}
Mehdi Neshat and Mehdi Yaghobi.
\newblock Feshdd: fuzzy expert system for hepatitis b diseases diagnosis.
\newblock In {\em 2009 Fifth International Conference on Soft Computing,
  Computing with Words and Perceptions in System Analysis, Decision and
  Control}, 2009.

\bibitem{adeli2013new}
Mahdieh Adeli, Nooshin Bigdeli, and Karim Afshar.
\newblock New hybrid hepatitis diagnosis system based on genetic algorithm and
  adaptive network fuzzy inference system.
\newblock In {\em 2013 21st Iranian conference on electrical engineering
  (ICEE)}, pages 1--6. IEEE, 2013.

\bibitem{ahmad2018intelligent}
Waheed Ahmad, Ayaz Ahmad, Amjad Iqbal, Muhammad Hamayun, Anwar Hussain, Gauhar
  Rehman, Salman Khan, Ubaid~Ullah Khan, Dawar Khan, and Lican Huang.
\newblock Intelligent hepatitis diagnosis using adaptive neuro-fuzzy inference
  system and information gain method.
\newblock {\em Soft Computing}, pages 1--8, 2018.

\bibitem{ahmad2019automated}
Gulzar Ahmad, Muhammad~Adnan Khan, Sagheer Abbas, Atifa Athar, Bilal~Shoaib
  Khan, and Muhammad~Shoukat Aslam.
\newblock Automated diagnosis of hepatitis b using multilayer mamdani fuzzy
  inference system.
\newblock {\em Journal of healthcare engineering}, 2019, 2019.

\bibitem{jang1991fuzzy}
Jyh-Shing~Roger Jang et~al.
\newblock Fuzzy modeling using generalized neural networks and kalman filter
  algorithm.
\newblock In {\em AAAI}, volume~91, pages 762--767, 1991.

\bibitem{tahmasebi2012hybrid}
Pejman Tahmasebi and Ardeshir Hezarkhani.
\newblock A hybrid neural networks-fuzzy logic-genetic algorithm for grade
  estimation.
\newblock {\em Computers \& geosciences}, 42:18--27, 2012.

\bibitem{karaboga2018adaptive}
Dervis Karaboga and Ebubekir Kaya.
\newblock Adaptive network based fuzzy inference system (anfis) training
  approaches: a comprehensive survey.
\newblock {\em Artificial Intelligence Review}, pages 1--31, 2018.

\bibitem{nilashi2019predictive}
Mehrbakhsh Nilashi, Hossein Ahmadi, Leila Shahmoradi, Othman Ibrahim, and Elnaz
  Akbari.
\newblock A predictive method for hepatitis disease diagnosis using ensembles
  of neuro-fuzzy technique.
\newblock {\em Journal of infection and public health}, 12(1):13--20, 2019.

\bibitem{trelea2003particle}
Ioan~Cristian Trelea.
\newblock The particle swarm optimization algorithm: convergence analysis and
  parameter selection.
\newblock {\em Information processing letters}, 85(6):317--325, 2003.

\bibitem{zhan2009adaptive}
Zhi-Hui Zhan, Jun Zhang, Yun Li, and Henry Shu-Hung Chung.
\newblock Adaptive particle swarm optimization.
\newblock {\em IEEE Transactions on Systems, Man, and Cybernetics, Part B
  (Cybernetics)}, 39(6):1362--1381, 2009.

\bibitem{neshat2010aipso}
Mehdi Neshat and Masoud Rezaei.
\newblock Aipso: adaptive informed particle swarm optimization.
\newblock In {\em 2010 5th IEEE International Conference Intelligent Systems},
  pages 438--443. IEEE, 2010.

\bibitem{neshat2012new}
Mehdi Neshat, Mehdi Sargolzaei, Azra Masoumi, and Adel Najaran.
\newblock A new kind of pso: Predator particle swarm optimization.
\newblock {\em International Journal on Smart Sensing \& Intelligent Systems},
  5(2), 2012.

\bibitem{hu2012adaptive}
Mengqi Hu, Teresa Wu, and Jeffery~D Weir.
\newblock An adaptive particle swarm optimization with multiple adaptive
  methods.
\newblock {\em IEEE Transactions on Evolutionary Computation}, 17(5):705--720,
  2012.

\bibitem{neshat2013faipso}
Mehdi Neshat.
\newblock Faipso: fuzzy adaptive informed particle swarm optimization.
\newblock {\em Neural Computing and Applications}, 23(1):95--116, 2013.

\bibitem{lin2019adaptive}
Anping Lin, Wei Sun, Hongshan Yu, Guohua Wu, and Hongwei Tang.
\newblock Adaptive comprehensive learning particle swarm optimization with
  cooperative archive.
\newblock {\em Applied Soft Computing}, 77:533--546, 2019.

\bibitem{naka2003hybrid}
Shigenori Naka, Takamu Genji, Toshiki Yura, and Yoshikazu Fukuyama.
\newblock A hybrid particle swarm optimization for distribution state
  estimation.
\newblock {\em IEEE Transactions on Power systems}, 18(1):60--68, 2003.

\bibitem{zahara2009hybrid}
Erwie Zahara and Yi-Tung Kao.
\newblock Hybrid nelder--mead simplex search and particle swarm optimization
  for constrained engineering design problems.
\newblock {\em Expert Systems with Applications}, 36(2):3880--3886, 2009.

\bibitem{wang2018hybrid}
Feng Wang, Heng Zhang, Kangshun Li, Zhiyi Lin, Jun Yang, and Xiao-Liang Shen.
\newblock A hybrid particle swarm optimization algorithm using adaptive
  learning strategy.
\newblock {\em Information Sciences}, 436:162--177, 2018.

\bibitem{DBLP:conf/gecco/Auger09}
Anne Auger.
\newblock Benchmarking the {(1+1)} evolution strategy with one-fifth success
  rule on the {BBOB-2009} function testbed.
\newblock In {\em Genetic and Evolutionary Computation Conference, {GECCO}
  2009, Proceedings, Montreal, Qu{\'{e}}bec, Canada, July 8-12, 2009, Companion
  Material}, pages 2447--2452, 2009.

\bibitem{muthukaruppan2012hybrid}
S~Muthukaruppan and Meng~Joo Er.
\newblock A hybrid particle swarm optimization based fuzzy expert system for
  the diagnosis of coronary artery disease.
\newblock {\em Expert Systems with Applications}, 39(14):11657--11665, 2012.

\end{thebibliography}

\end{document}